%% file: main.tex
\documentclass[pmlr]{jmlr}% new name PMLR (Proceedings of Machine Learning)

\usepackage{longtable}% for long tables

 \usepackage{booktabs}
 
\usepackage[load-configurations=version-1]{siunitx} % newer version
\usepackage{xcolor}
\usepackage{soul}

\usepackage[utf8]{inputenc}
\usepackage{enumitem}

\usepackage{algorithm,algpseudocode,float}

\makeatletter
\def\set@curr@file#1{\def\@curr@file{#1}} %temp workaround for 2019 latex release
\makeatother

\theorembodyfont{\upshape}
\theoremheaderfont{\scshape}
\theorempostheader{:}
\theoremsep{\newline}

\jmlrvolume{182}
\jmlryear{2022}
\jmlrworkshop{Machine Learning for Healthcare}

\title[Deep Instance Nuclei Detection]{Weakly Supervised Deep Instance Nuclei Detection using Points Annotation in 3D Cardiovascular Immunofluorescent Images}

\author{\Name{Nazanin Moradinasab}
       \Email{nm4wu@virginia.edu}\\ 
       \addr Department of Engineering Systems and Environment\\
       University of Virginia\\
       Charlottesville, VA, USA 
       \AND
       \Name{Yash Sharma}
       \Email{ys5hd@virginia.edu}\\ 
       \addr Department of Pediatrics\\
       University of Virginia\\
       Charlottesville, VA, USA
       \AND
       \Name{Laura S. Shankman}
       \Email{lss4f@virginia.edu}\\ 
       \addr Laboratory of Dr. Gary Owens Cardiovascular Research Center\\
       University of Virginia\\
       Charlottesville, VA, USA
       \AND
       \Name{Gary K. Owens}
       \Email{gko@virginia.edu}\\ 
       \addr Department of Molecular Physiology and Biological Physics\\
       University of Virginia\\
       Charlottesville, VA, USA
       \AND
       \Name{Donald E. Brown}
       \Email{deb@virginia.edu}\\ 
       \addr School of Data Science\\
       University of Virginia\\
       Charlottesville, VA, USA} 

\editor{Editor's name}

\begin{document}

\maketitle

\begin{abstract}
%   Summary of the article.  Be sure to highlight how the work
%   contributes to our understanding of machine learning and healthcare.
\input{Abstract}

\end{abstract}

\section{Introduction}
\input{Introduction}

\section{Related Work}
\input{Related_Work}

\section{Method}
\input{Methods}

\section{Results}

\input{Results_on_Real_Data}

\section{Discussion} 
\input{Discussion}

% {\color{red}

% For the submission, please do \emph{not} include the name of the
% institutions for any private data sources.  However, in the
% camera-ready, you may include identifying information about the
% institution as well as should include any relevant IRB approval
% statements. also Acknowledgement

% % ACKNOWLEDGEMENTS ONLY GO IN THE CAMERA-READY, NOT THE SUBMISSION
% % \acks{Many thanks to all collaborators and funders!}
% }
\bibliography{refs.bib}

% \appendix
% \section*{Appendix A.}

\end{document}

%% file: Abstract.tex
  %purpose
  %problem
  %method
  %results
  %conclusion
  
Two major causes of death in the United States and worldwide are stroke and myocardial infarction. The underlying cause of both is thrombi released from ruptured or eroded unstable atherosclerotic plaques that occlude vessels in the heart (myocardial infarction) or the brain (stroke). Clinical studies show that plaque composition plays a more important role than lesion size in plaque rupture or erosion events. To determine the plaque composition, various cell types in 3D cardiovascular immunofluorescent images of plaque lesions are counted. However, counting these cells manually is expensive, time-consuming, and prone to human error. These challenges of manual counting motivate the need for an automated approach to localize and count the cells in images. The purpose of this study is to develop an automatic approach to accurately detect and count cells in 3D immunofluorescent images with minimal annotation effort. In this study, we used a weakly supervised learning approach to train the HoVer-Net segmentation model using point annotations to detect nuclei in fluorescent images. The advantage of using point annotations is that they require less effort as opposed to pixel-wise annotation. 
To train the HoVer-Net model using point annotations, we adopted a popularly used cluster labeling approach to transform point annotations into accurate binary masks of cell nuclei. Traditionally, these approaches have generated binary masks from point annotations, leaving a region around the object unlabeled (which is typically ignored during model training). However, these areas may contain important information that helps determine the boundary between cells. Therefore, we used the entropy minimization loss function in these areas to encourage the model to output more confident predictions on the unlabeled areas. Our comparison studies indicate that the HoVer-Net model trained using our weakly supervised learning approach outperforms baseline methods on the cardiovascular dataset. In addition, we evaluated and compared the performance of the trained HoVer-Net model to other methods on another cardiovascular dataset, which also utilizes DAPI to identify nuclei, but is from a different mouse model stained and imaged independently from the first cardiovascular dataset. The comparison results show the high generalization capability of the HoVer-Net model trained using a weakly supervised learning approach and assessed with standard detection metrics.

%% file: Introduction.tex
Two major causes of death in the United States and worldwide are stroke and myocardial infarction (MI) (\citealt{virmani2000lessons}). The underlying cause of both is thrombi released from ruptured or eroded unstable atherosclerotic plaques that occlude vessels in the heart (MI) or the brain (stroke) (\citealt{biccard2018perioperative, rickard2016associations}). Unstable plaques are more prone to rupture or erosion, leading to possible MI or stroke. 
Human morphological studies have shown that the critical factor of plaque stability is plaque composition rather than lesion size (\citealt{libby2012inflammation, pasterkamp2017temporal}).
\cite{virmani2000lessons} have extensively studied the composition of human lesions and established that lesions with a thin extracellular matrix (ECM)- rich protective fibrous cap and a predominance of CD68+ relative to ACTA2+ cells, presumed to be macrophages (M$\Phi$) and smooth muscle cells (SMC), respectively, are prone to plaque rupture (\citealt{davies1993risk}). Another study showed that loss of endothelial cells (EC) overlying lesions, and increased CD31+ ACTA2+ cells assumed to be EC that have undergone EC to mesenchymal transition (EndoMT), are prone to erosion. However, lineage tracing studies in \cite{adorno2021combining} highlighted that more than 80\% of SMCs in advanced mouse atherosclerotic lesions no longer had detectable ACTA2. Further, a subset of these cells expressed LGALS3, a marker that would have traditionally classified them as a (M$\Phi$).

Plaque composition can be determined by immunofluorescent staining of histological cross-sections from diseased vessels. These morphological studies are extremely valuable for understanding the underlying mechanisms for plaque rupture and determining the mechanisms that promote atherosclerotic plaque stability. However, to determine the plaque composition from the immunofluorescent images, we need to first accurately detect cells and assign them phenotypes based on co-expressed markers. This process requires hours of manual detection of the various cell types, which is slow, expensive, and prone to human error. The challenges of manual cell counting indicate a need for automated image processing to localize and count various cell types in order to find their distribution in fluorescent microscopy images, and thereby classify lesions as stable or unstable.  
Several challenges arise when designing automated image analysis, such as the heterogeneity of cell types (shape and size), autofluorescent signal from the tissue, and low image contrast (\citealt{xing2013automatic}). Moreover, automatic localization and counting of various cell types in 3D immunofluorescent images have the added challenge of overlapping cells and cellularly dense regions that are difficult to count. In addition, there can be variability in the depth of imaging and thickness of the sample tissue itself.      

Recently, modern deep learning-based nuclei segmentation approaches (\citealt{graham2019hover, kumar2017dataset, naylor2017nuclei, mahmood2019deep, qu2019joint}) have become popular over traditional methods (\citealt{arteta2016detecting}) to quantify the histopathology and fluorescent microscopy images. However, these neural networks are usually categorized as fully supervised approaches that require a large amount of pixel-wise annotated data for training. Collecting pixel-wise annotated data is expensive, time-consuming, and difficult because it requires classification of every pixel in an image and is near impossible to perform on 3D images. Alternatively, adapting weak-annotation methods such as labeling each nucleus with a point reduces the burden of pixel-wise annotations. 
Several studies (\citealt{qu2019weakly, chamanzar2020weakly, nishimura2019weakly}) have tried to address training neural networks based on point annotations but focus on the nuclei segmentation problem using point annotations. In these approaches, as point annotations alone are not sufficient to train a neural network model, the authors take advantage of the original images and the shape of nuclei, among others, to get extra information to train the model. None of these studies pay attention to the nucleus' boundary (\citealt{tian2020weakly}), even though it plays a key role to separate clustered nuclei. 

\cite{graham2019hover} proposed the HoVer-Net approach for Simultaneous Segmentation and Classification of Nuclei in Multi-Tissue Histology Images. One of the key advantages of this method is that it utilizes the horizontal and vertical distances of nuclear pixels to their center of mass to separate clustered nuclei. These horizontal and vertical distances can inform the model where to separate neighboring nuclei, resulting in increased performance in situations where the image contains nuclei dense regions. Another noticeable advantage of the HoVer-Net model is its generalization capability. The authors assessed the generalization capability of the HoVer-Net model from two aspects, including 1) its performance over samples originating from different organs (variation in nuclei shapes) and 2) its performance over samples originating from the same organ but different sources (variation in staining). In both situations, the results indicate the HoVer-Net model can successfully generalize to unseen data. 

In this paper, our aim is to detect and localize nuclei in 3D immunofluorescent images that contain regions of crowded nuclei. Fully annotated segmentation maps at pixel level are required for model training. However, this process is expensive, especially in 3D dimensions. To reduce the burden of annotation, we asked the researchers to perform point annotations (i.e., marking the center of each nucleus with a point in two dimensions, including x and y on the 3D image). Since plaque stability assessment focuses on cell composition within the atherosclerotic lesion, the researchers were only required to mark nuclei within this region of interest (ROI). Due to the variability in the number of images in the z-plane, we opted to collapse the 3D image into a 2D image using a Maximum Intensity Projection (MIP) (\citealt{napel1992ct}). While MIP improves the computation of the model, it also increases the number of nuclei that may touch or overlap (\citealt{ho2020sphere}). HoVer-Net excels in detecting neighboring nuclei but needs pixel-wise labels to train the model. However, due to having only access to the point annotations for our dataset, we use the weakly supervised learning approach to train the HoVer-Net model. We generated the pixel-level labels using the point annotation approach proposed in \cite{qu2019weakly}. In \cite{qu2019weakly}, the coarse pixel-wise labels are generated and refined using k-mean clustering and the Voronoi diagram, respectively. However, one of the challenges of utilizing these coarse pixel-wise labels is that the boundaries of the nuclei are not clear, especially in z-stack images, i.e., the boundaries of the nuclei are dimmer in comparison to the centers of the nuclei. In this paper, we consider the boundaries of nuclei as unlabeled areas, and we label them neither as nuclei nor as a background. We adapt the semi-supervised learning approach, and we only use entropy minimization loss over the boundaries of the nuclei as unlabeled areas.   

\subsection*{Generalizable Insights about Machine Learning in the Context of Healthcare}

In this paper, to detect nuclei in 3D immunofluorescent images, we adapt the weakly-supervised learning approach to train the HoVer-Net model using point annotations. The main contributions of this work are listed as follows:
\begin{itemize}
   \item Adopting HoVer-Net model to detect nuclei on the 3D immunofluorescent images using point annotations (new application).
   \item Adapting semi-supervised learning approach to train model over the boundaries of nuclei.
   \item We show that the HoVer-Net model outperforms the current weakly supervised learning approaches in detecting clusters of nuclei on the 3D immunofluorescent images. 
   \item We show that the HoVer-Net model requires no modification for training in a weakly supervised learning setup
   \item We show the generalization capability of the HoVer-Net model in comparison with the current weakly supervised learning approaches in detecting clusters of nuclei on the 3D immunofluorescent images.
   
\end{itemize}

%% file: Related_Work.tex
Fully supervised training is the most favored algorithm for semantic segmentation. However, these approaches require expensive pixel-level annotations for training. Hence, for reducing the annotation burden, multiple weak supervised learning approaches incorporating different types of weak labels for training have been proposed, such as image-level (\citealt{pinheiro2015image, chang2020weakly}), point-level (\citealt{bearman2015feifei, tian2020weakly}, \citealt{yoo2019pseudoedgenet}), bounding box (\citealt{papandreou2015weakly, dai2015boxsup}, \citealt{khoreva2017simple}), and scribble (\citealt{lin2016scribblesup, vernaza2017learning}). Compared to pixel-level annotations, these approaches use spatially less informative annotations for training. 

Further, depending on the domain of images, each domain presents a unique set of problems for weakly supervised segmentation. \cite{chan2021comprehensive} highlighted the challenges of extending natural image approaches to other domains such as medical imaging and satellite imaging. Medical imaging data contains finer-grained objects, leading to difficulties in accurately detecting boundaries and class co-occurrence compared to natural image data that contain coarse-grained visual information. Moreover in medical imaging data, accurate detection and segmentation of nuclei in images plays a critical role in the diagnosis and prognosis of patients. Hence, we used point-level annotation to accurately detect and segment the nuclei while reducing the annotation burden. Below we provide an overview of some of the recently proposed 1) Nuclei instance segmentation techniques and 2) Weakly Supervised Image Segmentation using Point annotation.

\subsection{Nuclei Instance Segmentation}

Due to the small size of nuclei and their overlapping structures, nuclei instance segmentation is a challenging task. As a result, different strategies for separating nuclear boundaries have been proposed in the literature. \cite{kumar2017dataset} incorporated boundary pixels with nuclei and background for the segmentation model training and performed anisotropic region growing as a post-processing step. \cite{kang2019nuclei} further extended the three-class segmentation approach by using it as an intermediate task for estimating coarse boundaries followed by fine-grained segmentation. \cite{naylor2017nuclei} formulated the segmentation problem as a regression task of the distance map for separating the touching or overlapping nuclei. \cite{schmidt2018cell} used star-convex polygons for localizing cell nuclei. Another branch of approach that has shown promising results uses auxiliary task learning to separate overlapping nuclei. \cite{chen2017dcan} proposed a deep contour-aware network integrating instance appearance and contour information into a multi-task learning framework and a weighted auxiliary classifier to address the vanishing gradient problem. \cite{oda2018besnet}, in their Boundary-Enhanced Segmentation Network, added another decoding path in the U-Net architecture for enhancing the boundaries of cells. \cite{liu2019nuclei} designed a dual-branch segmentation model integrating the auxiliary semantic segmentation branch with the instance segmentation branch via a feature fusion mechanism. \cite{zhou2019cia} proposed a Contour-aware informative aggregation network aggregating the spatial and texture dependencies of nuclei and contour in the decoder's bi-directional feature aggregation module. Hover-Net, one of the popular methods for instance segmentation and classification, used horizontal and vertical distance maps to the nuclear center with segmentation and classification maps for learning. Most recently, \cite{he2021cdnet} learned the spatial relationship between nucleus pixels via the centripetal direction feature. These direction features were then used to separate instances.

\subsection{Weakly Supervised Image Segmentation using Point annotation}

Since \cite{bearman2015feifei} proposed point annotations for semantic segmentation and established it as an effective strategy for object detection and counting tasks, it has been extended to other domains, including medical imaging and nuclei segmentation. \cite{zhou2018sfcn} designed architecture with sibling branches for cell nuclei detection and classification tasks and trained them using centroid point annotations. \cite{yoo2019pseudoedgenet} introduced an auxiliary task, Pseudoegnet, for accurately detecting nuclei boundaries without edge annotations. \cite{nishimura2019weakly} used contribution pixel analysis in the centroid detection network for instance segmentation. They used guided backpropagation focusing on particular regions for determining the contributing pixels for a predicted centroid. \cite{qu2019weakly} generated the Voronoi label and cluster label from the point label and used them to train the U-Net model with CRF loss for segmentation. In the follow-up to this work, they tackled a more challenging scenario of partial point annotation and used a two-stage learning framework for nuclei segmentation (\citealt{qu2019joint}). In the first stage, they used self-training for generating nuclei annotation for the unlabeled region, followed by their weakly supervised segmentation module. \cite{chamanzar2020weakly} used Voronoi transformation, local pixel clustering, and repel encoding for generating pixel-level labels for U-Net training via a multi-task scheduler. \cite{tian2020weakly} proposed a coarse-to-fine two-staged training framework. In the first stage for generating coarse maps, they employed an iterative self-supervision strategy for generating high confidence point-distance maps along with Voronoi edge distance maps for training. Further, in the second stage, they refined predictions by incorporating contour-sensitive constraints. 

%% file: Methods.tex
\begin{figure}[t]
  \centering 
  \includegraphics[width=5in]{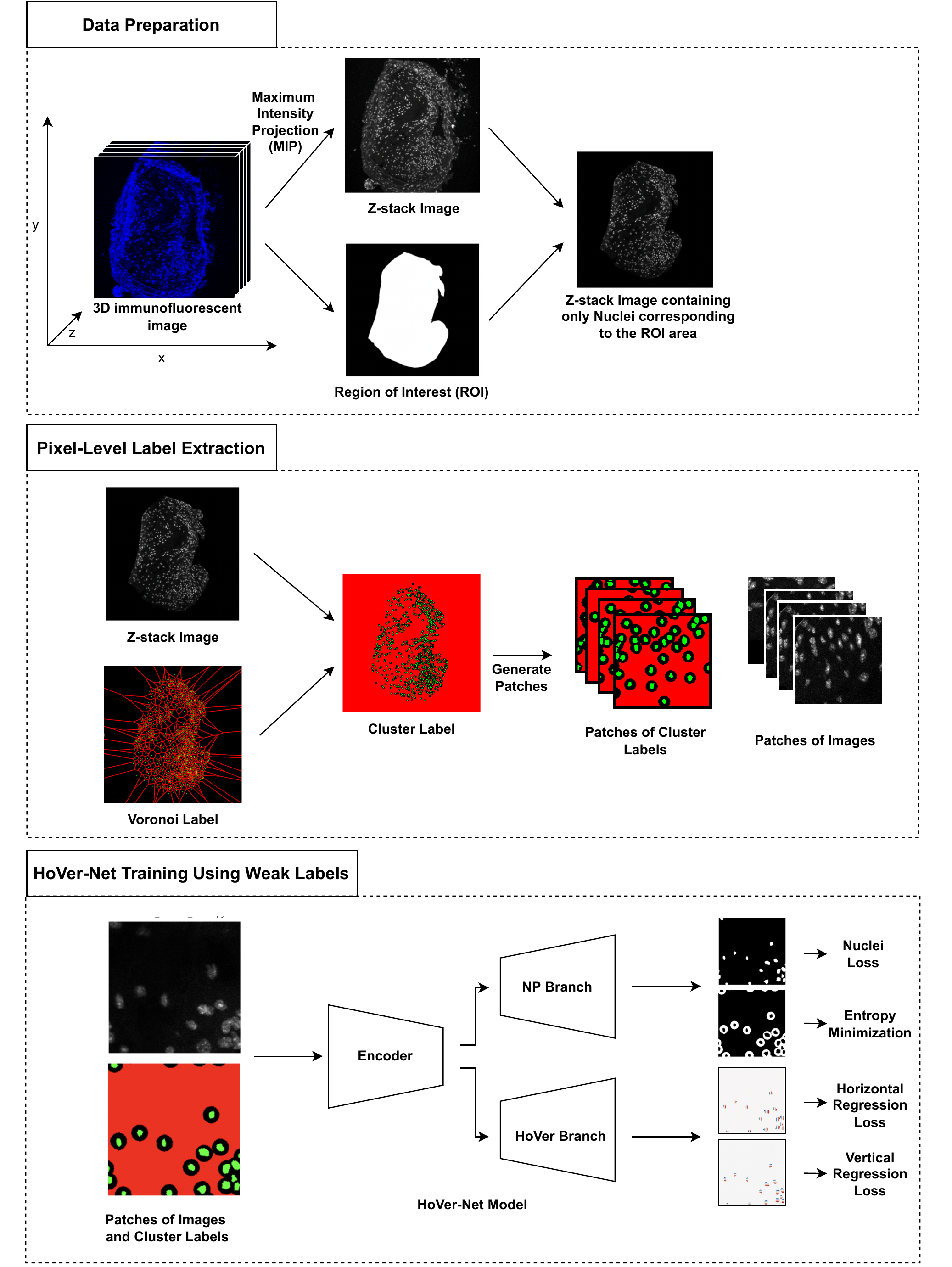} 
  \caption{Diagram of Proposed Approach}
  \label{fig:approach} 
\end{figure}

In this section, we discuss the details of the proposed approach to detect nuclei on 3D immunofluorescent images, including data preparation, pixel-level label extraction, and training algorithm. The schematic diagram of the proposed approach is presented in Figure \ref{fig:approach}.

\subsection{Data preparation}
In this study, we use two independent Cardiovascular datasets (i.e., D1 and D2) containing 3D immunofluorescent images and their corresponding point annotations. These datasets are both from the same region of the brachiocephalic artery (BCA) but collected from different mouse models and prepared/imaged independently of each other. 
The size of D1 and D2 is equal to 10 and 19, respectively. We split D1 into train and test sets with sizes 8 and 2. We train the model over the train set and evaluate it over the test set and D2. In the first step, we convert these 3D images into 2D images by using the Maximum Intensity Projection function (\citealt{napel1992ct}), as shown in Figure \ref{fig:approach}. Next, since the cells were only annotated within specified ROI, we zeroed the areas of the obtained z-stack images outside of the ROI to prevent unlabeled cells from affecting training and evaluation. Because these immunofluorescent images are high-resolution images and due to high computational complexity, it is not possible to train the CNN algorithm over the entire image at once, therefore we split them into smaller patches. We extract $256\times256$ pixel patches from images with 10\% overlap. The 10\% overlap was used to make sure all the cells were preserved. The total number of patches in the train and test sets are equal to 383 and 97, respectively.

\subsection{Pixel-level label extraction}

We use the Hover-Net model for segmentation due to its strong generalizability and instance detection performance. As we can not directly use the point-level labels for training the Hover-Net model, we used the cluster label approach proposed in \cite{qu2019weakly} to extract pixel-level labels from point annotation via information obtained from the shape of nuclei in the original image. The pseudo-code for creating the cluster label is depicted in Algorithm \ref{pseudocode}, where K and N are the number of clusters and images, respectively. The number of the clusters is set to three (K=3), indicating nuclei, background, and ignored class. In the first two steps of this algorithm, a Voronoi label and distance map for each z-stack image are computed. The distance map is generated by calculating the distance of each pixel to the closest nuclear point. Then, the k-mean clustering is applied to the combination of the distance map and RGB values of the original image. Concurrently Voronoi diagram is used to partition a plane into $n$ regions based on $n$ seed points. These regions have three characteristics: i) they are convex polygons, ii) each region contains exactly one seed point, and iii) every point in a given region is closer to its seed point than other seed points. As depicted in Figure \ref{fig:approach}, we refine the cluster label using the Voronoi label to improve the separation of clustered nuclei, resulting in more accurate masks with which to train the model. We do not use the Voronoi labels independently to train the HoVer-Net model because this model is capable of detecting the cluster of nuclei separately using the horizontal and vertical maps. 

\begin{algorithm}[H]
\SetAlgoLined
\SetKwInOut{Input}{input}

Initialize K, \\
\Input{Original z-stack images, Point-level labels}
\SetKwInOut{Parameter}{parameter}
 \For{$i = 1: N$}{
 \begin{enumerate}[noitemsep]
 \vspace{-0.2cm}\item Generate the Voronoi label for the image i \\
 \vspace{-0.5cm}\item Generate the distance map for the image i $(D_{i})$\\
 \vspace{-0.5cm}\item Clip the values in the distance map by truncating large values to 20\\
 \vspace{-0.5cm}\item Normalize the RGB values to have the same range as the distance values\\
 \vspace{-0.5cm}\item Combine the distance map with the RGB channels to create the feature map\\
 \vspace{-0.5cm}\item Apply k-mean clustering on the feature map with the loss function $\underset{S}{\arg\max}\underset{j=1}\Sigma^{k}\underset{x\in S_{j}}\Sigma||f_{x}-c_{j}||$\\
 \vspace{-0.3cm}\item Identify the cluster with the maximum overlap with the point labels and designate it nuclei\\
 \vspace{-0.5cm}\item 
 Identify the cluster with the maximum overlap with the dilated point labels and designate it background\\
 \vspace{-0.5cm}\item The remaining class is designated as the ignored class\\
 \vspace{-0.5cm}\item Assign the corresponding cluster label to each pixel of the image i\\
 \vspace{-0.5cm}\item Refine the cluster label by using the Voronoi label
 \end{enumerate}
 }
  
\caption{Generating the cluster labels by using point annotations and original images \cite{qu2019weakly}}
\label{pseudocode}
\end{algorithm}

\begin{figure}[t]
  \centering 
  \includegraphics[width=3.5in]{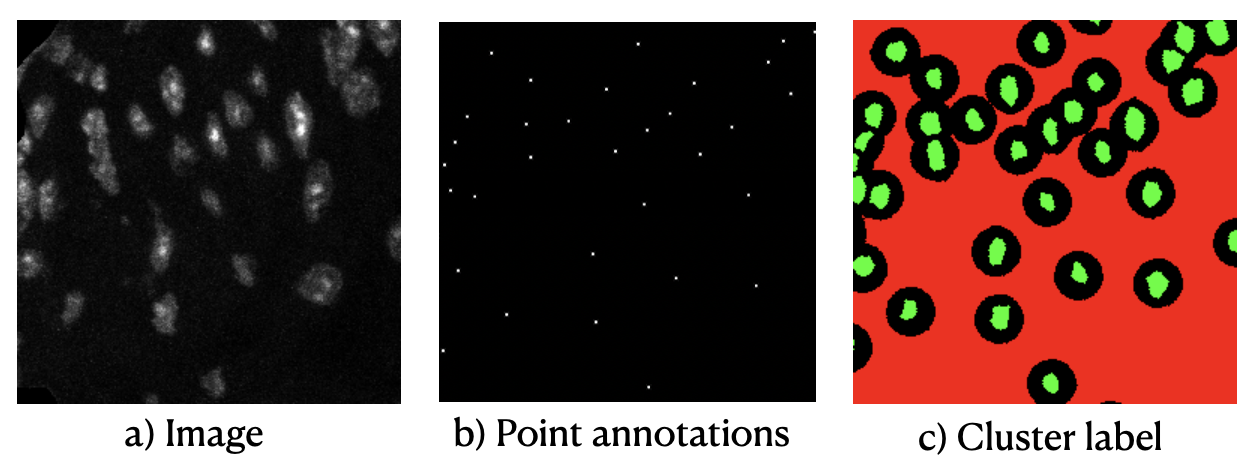} 
  \caption{(a) Original image, (b) Point annotation, (c) Cluster label which is refined using Voronoi diagram}
  \label{fig:label} 
\end{figure}

In Qu's approach (\citealt{qu2019weakly}), the authors do not train their model on the ignored areas because these ignored areas are often located around the nuclear boundaries, and it is difficult to assign the appropriate label to them (i.e., Background or Nuclei). However, these areas may contain important information to improve nuclei detection. Therefore, we adapted a semi-supervised learning approach and used entropy minimization loss function in these unlabeled areas to train the model over them without requiring labels. The entropy minimization loss encourages the model to output confident predictions over these unlabeled areas.  
In Figure \ref{fig:label}, the final cluster label (Fig. \ref{fig:label}(c)) is presented for original image (Fig. \ref{fig:label}(a)). In this figure, the green, red and black colors indicate the nuclei, background, and unlabeled areas, respectively. 

\subsection{Algorithm}

As depicted in Figure \ref{fig:approach}, we adopted the HoVer-Net (\citealt{graham2019hover}) model and trained it using a weakly supervised learning approach to detect nuclei in z-stack fluorescent images. The HoVer-Net model can simultaneously obtain accurate nuclear instance segmentation and classification by using three branches: (i) nuclear pixel (NP) branch; (ii) HoVer branch and (iii) nuclear classification (NC) branch. The NP branch predicts the label of each pixel as nuclei or background. And the HoVer branch is used to separate the neighboring nuclei by predicting the horizontal and vertical distances of nuclear pixels to their centers of mass. Finally, the NC branch predicts the type of each nucleus. In this report, we focus on nuclei detection and thereby only consider the NP and HoVer branches. The HoVer-Net loss functions consist of 1) the cross-entropy loss, 2) dice loss, and 3) the regression loss. The first two losses are computed at the output of the NP branch, and the last loss is obtained at the HoVer branch's output. 

\begin{figure}[t]
  \centering 
  \includegraphics[width=4.5in]{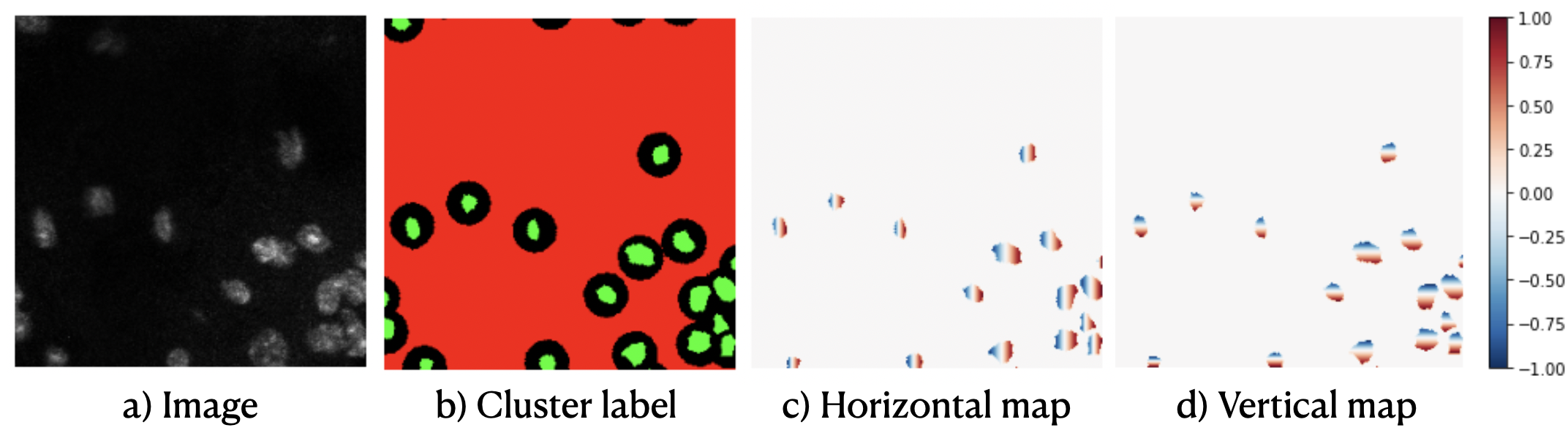} 
  \caption{(a) Original image, (b) Cluster label,  (c)  Horizontal map, (d) Vertical map}
  \label{fig:hovernet_label} 
\end{figure}

To train the HoVer-Net model using point annotations, we need to create three masks involving cluster labels, and horizontal and vertical maps, as shown in Figure \ref{fig:hovernet_label}. First, we need to generate the pixel-level labels using point labels and original images as described in the previous section to supervise the NP branch (Figure \ref{fig:hovernet_label}(b)). These generated pixel-level labels are not perfect, and they contain three areas: green, red, and black, corresponding to nuclei, background, and unlabeled areas, respectively. The reason for having the unlabeled pixels is that, as mentioned before, the boundaries of the nuclei are usually dimmer in comparison to the centers of it in z-stack images, which makes it harder to assign labels to the nuclei boundaries using the clustering methods. In general, we have labeled (green and red section) and unlabeled (black section) areas in each generated pixel-level label. 
To train the HoVer-Net branch over these generated pixel-level labels, we make a small modification to the calculation of the loss function at the output of the NP branch. We adapt a semi-supervised learning approach used in conditions where we are dealing with labeled and unlabeled data. A common underlying assumption in semi-supervised learning is that the classifier's decision boundary should pass the low-density area. One technique to encourage this is to enforce the classifier to produce low-entropy predictions on the unlabeled data (\citealt{berthelot2019mixmatch}). To adapt this method to our problem, we calculate the cross-entropy and dice losses at the labeled area (green and red) and entropy minimization at an unlabeled area (black) at the output of the NP branch. The entropy minimization loss function is as follows:
\begin{equation}
  H(p) = -\Sigma_{i=1}^{m} p_{i}log p_{i}  
\end{equation}

It should be noted that to train the Hover branch, we need to generate the Horizontal and Vertical maps for all images. Figure \ref{fig:hovernet_label}(c-d) indicates one sample of these mask generated for original image (Figure\ref{fig:hovernet_label}(a)).

%% file: Results_on_Real_Data.tex
In this section, we discuss the details of algorithm implementations and evaluations and compare our proposed approach with recently proposed weakly-supervised training approaches via point annotations.

\subsection{Evaluation Approach/Study Design} 

\subsubsection{Metrics}
Our goal is to accurately detect and quantify cells on the images. To evaluate the model's performance, we need to compare the predicted and true cells. We use the popular detection metrics, including precision (P), recall (R), and F1 score, which are defined as follows:  

\begin{flalign}
    &P=\frac{TP}{TP+FP}\nonumber\\
    &R=\frac{TP}{TP+FN}\\
    &F1=\frac{2TP}{2TP+FP+FN}\nonumber
\end{flalign}

Where, TP, FP, and FN are the number of true positives, false positives, and false negatives, respectively. If the distance between the center of the detected nuclei and the closest point annotation is at most $r_{nuc}$, it is TP, otherwise, FP. If there are more than one nuclei in the $r_{nuc}$ distance of the given point annotation, the closest one is TP and the rest FP. $r_{nuc}$ is defined as the rough average of the nuclear radius and it is usually calculated using the validation set.  

\subsubsection{Datasets}

\begin{enumerate}
    \item  Two unique mouse models were used to generate advanced atherosclerotic lesion images contained in \textbf{Cardiovascular dataset 1 (D1)} and \textbf{Cardiovascular dataset 2 (D2)}. \textbf{Cardiovascular dataset 1 (D1)} consisted of mice lacking the leptin receptor (\textit{ob/ob} mice) causing them to become hyperphagic. These mice were also injected with a mutPCSK9 adenovirus to induce hypercholesterolemic when fed a high-fat Western diet for 18 weeks. D1 contains ten 3D immunofluorescent images with multiple channels created by antibody staining and scanning confocal microscopy on a LSM880 microscope. Because our goal is to detect cells in these images, we only use the channel corresponding to nuclei (DAPI). These images averaged 10 z-slices but were not consistent in number. Experts annotate images (i.e. marking the center of each nucleus with a point) using FIJI within the ROI. We then split this dataset into training (8 images) and test sets (2 images). \textbf{Cardiovascular dataset 2 (D2)} contains atherosclerotic lesions from \textit{Apoe\textsuperscript{-/-}} mice, a commonly used genetic background that makes mice susceptible to atherosclerosis when placed on a high-fat Western diet. D2  consists of nineteen 3D immunofluorescent images using the same nuclear marker (DAPI) as the previous dataset, but prepared and imaged independently. We use this dataset to evaluate and compare the generalization capability of the HoVer-Net model as well as other approaches. 
    \item \textbf{Multi-Organ (MO) dataset:} This is a public dataset containing 30 images of tumors of different organs released by \cite{kumar2017dataset}. The size of these images is $1000\times1000$, and they are collected from several patients at multiple hospitals. As this dataset contains images from different organs and cancer types, it has high variability. The full annotations for this dataset are available, and we generate the point annotations using them. The size of training, validation, and test sets are 12, 4, and 14, respectively. Note, the same image split was used in \cite{kumar2017dataset} and \cite{qu2019weakly}. 
\end{enumerate}

\subsubsection{Implementation and Training Details}

All models are developed with the open-source software library PyTorch version 1.8.0. The details of the implementation and hyperparameter values of each method are described as follows: 

\begin{figure}[t]
  \centering 
  \includegraphics[width=4.5in]{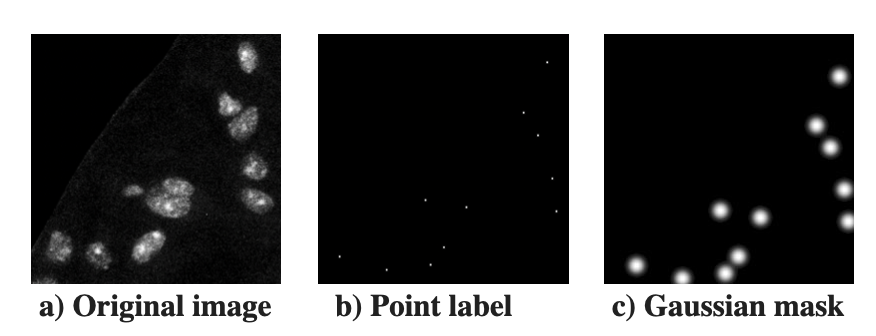} 
  \caption{(a) Original image, (b) Point annotation,  (c) Gaussian mask}
  \label{fig:Gaussian_mask} 
\end{figure}

\textbf{ResUnet34}: Pretrained Resnet34 is used in this study, and gaussian masks are generated using the Gaussian kernel with $\sigma=5$ to supervise the model. An example of generated gaussian masks using point annotation is shown in Figure \ref{fig:Gaussian_mask}. The intuition behind using a Gaussian mask is that it considers a high density around the point labels that is smoothly decreasing as we move away from point labels. These masks with blurry boundaries around the point labels are a good fit for detecting and localizing the nuclei, as the nuclei's boundaries are not clear while dealing with point labels. The learning rate is set to be 2e-4, and the model is trained using the regression loss function for 100 epochs. 

\textbf{HoVer-Net}: The HoVer-Net model is initialized with pre-trained weights on the ImageNet dataset (\citealt{deng2009imagenet}). The training process of the HoVer-Net model has two steps: first, we train only the decoder for 100 epochs and then fine-tune the whole model for another 100 epochs. The rest of the hyperparameters are set to values suggested by \cite{graham2019hover}. For training, entropy minimization loss is used for the unlabeled area, and dice loss, cross-entropy loss, and regression loss are used for the labeled area. The weights for all the loss functions are set to be 1 except that of entropy minimization, which is 0.5. We use two different strategies to train the HoVer-Net model using weekly supervised learning. In the first strategy indicated by HoVer-Net in table \ref{tab:dataset1}, we do not train the model on the unlabeled area (i.e. black area in Figure \ref{fig:hovernet_label}(b)). However, in the second strategy (i.e. HoVer-Net Ent), we use entropy minimization loss over the unlabeled areas. HoVer-Net Ent* indicates the best performance of the model after manually tuning the hyper-parameters.

\textbf{Qu's approach}: Most of the hyper-parameters are set to values suggested by \cite{qu2019weakly}. First, the model is trained for 100 epochs with a learning rate of 1e-4. Then, we fine-tune the model for 20 epochs using the dense CRF loss.

\subsection{Results on D1} 
The goal is to detect nuclei in 3D cardiovascular immunofluorescent images. In this section, first, we compare the performance of the HoVer-Net model trained using weakly supervised learning (WSL) with other baseline approaches in terms of precision, recall, and F1 score. In Table \ref{tab:baseline}, the HoVer-Net and the Qu’s models have been trained over the binary masks obtained using the cluster labeling approach (i.e. same labels but different architecture). The Qu's model has been trained over the Voronoi masks in addition to the cluster labeling masks. However, we train the ResUnet34 over Gaussian masks obtained using Gaussian filter, one of the popular methods to detect nuclei. All models are trained on the train set using point annotations and evaluated on the test set. Qu’s model use the Voronoi loss and clustering loss for training. Without the Voronoi loss function, Qu’s model doesn't work well in separating nuclei, as shown in Table 3 below. A common problem with the binary masks obtained using the cluster labeling approach is inaccurate object boundaries due to missing information. To handle this issue, the Dense CRF loss function is used in Qu's model, whereas the Hover-Net model can be trained for WSL without any fine-tuning. Table \ref{tab:baseline} indicates that the HoVer-Net model trained using imperfect binary masks obtained using the clustering labeling approach outperforms other methods. 

The binary masks obtained using the cluster labeling approach contain 3 areas: nuclei, background, and ignored regions.  \cite{qu2019weakly} proposed not to train the model over ignored regions as the labels of these regions are not clear. We follow this in Table \ref{tab:baseline} and do not train the Qu and HoVer-Net model over the ignored area. In our approach, we propose to use entropy minimization over the ignored region to encourage model to output high confident predictions as these areas may contain important cell boundary information. We extend the architecture of both HoVer-Net and Qu’ models by adding entropy minimization over the ignored regions. As shown in Table \ref{tab:extended}, the modified HoVer-Net (HoVer-Net Ent) and Qu’s model (Qu’s model Ent) have achieved better Recall and F1 and a competitive Precision. These results indicate that the ignored area contains important information and should not be ignored.

Due to the small size of nuclei and their overlapping structures, we need to use a specific strategy for separating nuclear boundaries, as was explained in the “Nuclei Instance Segmentation” Section. The HoVer-Net model used the horizontal and vertical maps, and the Qu’s model used the Voronoi labels for separating a cluster of nuclei. However, both of them are based on the ResUnet as an encoder. To analyze the effect of the entropy minimization over the ResUnet34, we implement the ablation study on the Qu’s model trained using the cross-entropy loss over both cluster and Voronoi labels (Table \ref{tab:ablation}). $\alpha$ is the balancing parameter for both cross-entropy losses, i.e. $\alpha$ = 0 means using only the cluster label loss, and $\alpha$ = 1 means using only the Voronoi label loss. Therefore, Qu’s model can be converted to the ResUnet model with the cluster label cross-entropy loss ($\alpha$=0). As per the results, the Entropy minimization method does not help to improve the model performance when we don't consider any nuclei separation strategy. And as shown in Table 3, with increasing the weight of the separation strategy (cross-entropy loss over the Voronoi label), we see the effect of the entropy minimization on the network performance.

\begin{figure}[t]
  \centering 
  \includegraphics[width=6in]{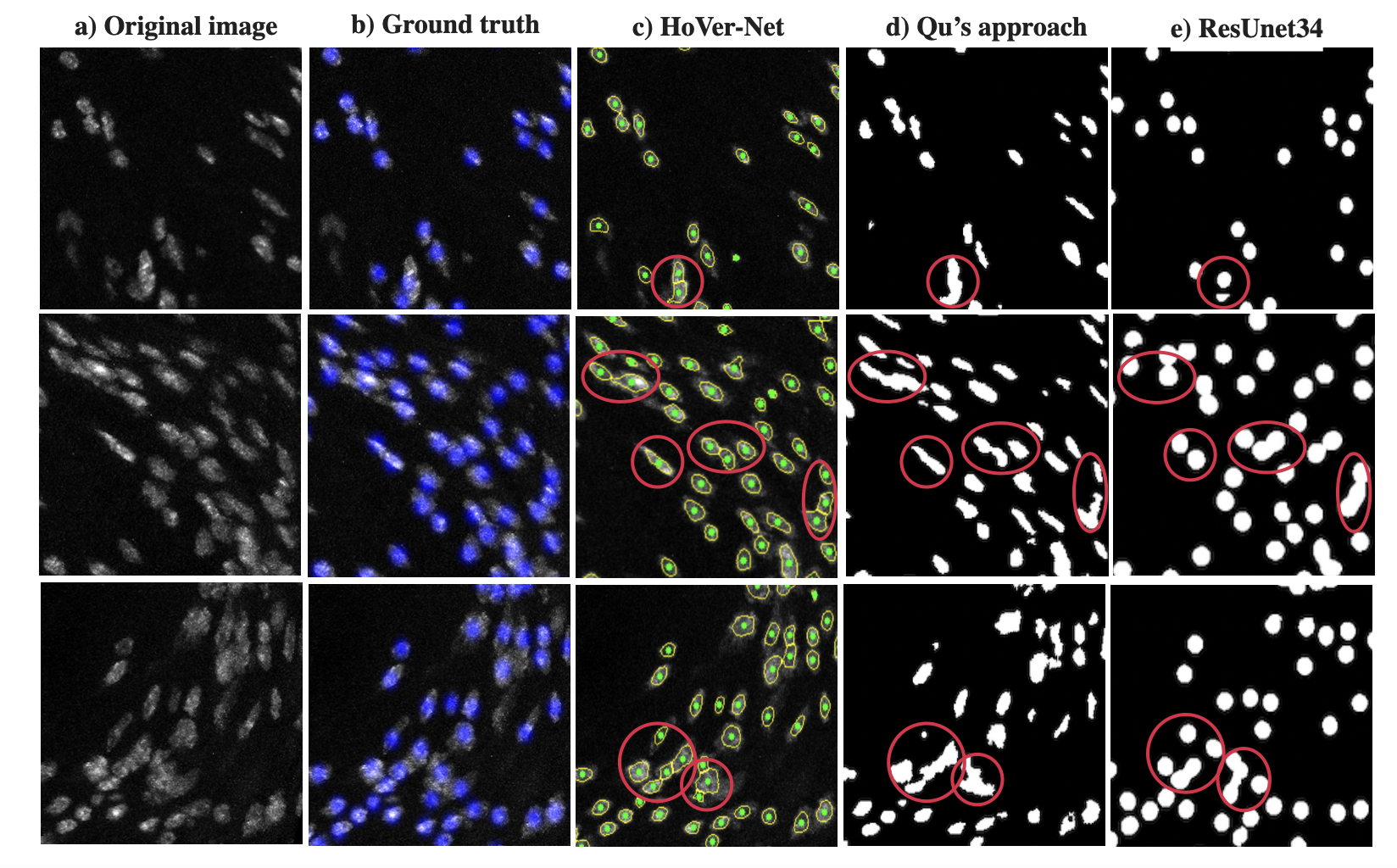} 
  \caption{Three samples of outputs generated by models (a) Original image, (b) Ground truth (i.e. point labels): the point labels are shown by larger circles to be more clear, (c) HoVer-Net: the detected boundary and center of the nuclei are shown by yellow and grean colors, respectively the, (d) Qu's model: the generated mask by model is shown by white color  of the model, (e) ResUnet:the generated mask by model is shown by white color of the model}
  \label{fig:result_sample} 
\end{figure}

For analyzing and comparing the three approaches mentioned above, the generated outputs for each are presented in Figure \ref{fig:result_sample}. As mentioned before, these z-stack immunofluorescent images contain crowded overlapped nuclei that are hard to detect separately by automatic algorithms. However, in comparison to two baseline approaches (i.e., Qu's model and ResUnet34), the HoVer-Net model can detect the neighboring instances better, as shown in Figure \ref{fig:result_sample}. Some of these close nuclei which the Qu's model and ResUnet34 failed to separate are shown by the red circle in Figure \ref{fig:result_sample}. 
In addition, it should be mentioned that one of the advantages of both Qu's model and the HoVer-Net model is the capability of capturing the nuclei's shape, as shown in Figure \ref{fig:result_sample}. Even though it does not affect the network's performance in detecting nuclei, it is essential when the goal is to classify the detected nuclei. One common approach that can be adopted for nuclei classification is to aggregate the pixel-level nuclear type predictions within each instance.

\begin{table}[t]
  \centering 
  \caption{Comparative experiments on the Results on D1}
  \begin{tabular}{lllll}
  \toprule
    \textbf{Model} & \textbf{Masks} & \textbf{Precision} & \textbf{Recall} & \textbf{F1} \\
    \midrule
    ResUnet34 & Gaussian masks & 0.8964 & 0.8551 & 0.8734 \\ 
    Qu's model & Clustering and Voronoi masks & 0.8824 & 0.7963 & 0.8371 \\ 
    HoVer-Net & Clustering masks & 0.914 & 0.867 & 0.888 \\
    
    \bottomrule
  \end{tabular}
  \label{tab:baseline} 
\end{table}

\begin{table}[t]
  \centering 
  \caption{Comparative experiments on the Results of Extended models on D1}
  \begin{tabular}{llll}
  \toprule
    \textbf{Architecture} & \textbf{Precision} & \textbf{Recall} & \textbf{F1} \\
    \midrule
    HoVer-Net & 0.9143 & 0.8673 & 0.8884\\ 
    HoVer-Net Ent & 0.9077 & 0.8897 & 0.8975 \\ 
    Qu’s model & 0.8824 & 0.7963 & 0.8371 \\
    Qu’s model Ent & 0.8526 & 0.8646 & 0.8585 \\
    
    \bottomrule
  \end{tabular}
  \label{tab:extended} 
\end{table}

\begin{table}[t]
  \centering 
  \caption{Ablation study to show the contribution of the Entropy minimization loss on the Qu’s model with different $\alpha$ values}
  \begin{tabular}{c|c|c|c|c|c|c}
  \toprule
    \textbf{Architecture} & \multicolumn{3}{|c|}{\textbf{Without Entropy minimization}} & \multicolumn{3}{|c}{\textbf{With Entropy minimization}} \\
    \cline{2-7}
    &\textbf{Precision} & \textbf{Recall} & \textbf{F1}&\textbf{Precision} & \textbf{Recall} & \textbf{F1}\\
    \midrule
    $\alpha: 0.0$ & 0.94 & 0.702 & 0.804 & 0.94 & 0.702 & 0.804\\ 
    $\alpha: 0.5$ & 0.911 & 0.737 & 0.815 & 0.896 & 0.829 & 0.861 \\ 
    $\alpha: 1.0$ & 0.882 & 0.796 & 0.837 &0.837 & 0.865 & 0.858\\
    \bottomrule
  \end{tabular}
  \label{tab:ablation} 
\end{table}

\subsection{Comparative Analysis of Generalization Capability} 
Achieving good performance on the unseen data is the ultimate goal of designing the automatic approach. \cite{graham2019hover} indicates that the HoVer-Net model trained using pixel-level labels has a generalization capability in terms of segmentation metrics. In this section, we compare the generalization capability of the HoVer-Net model for detection, trained using weakly supervised learning, with the baseline models on the unseen data. We evaluate the performance of model trained using D1 on the D2 in terms of precision, recall, and F1 score. As mentioned before, D2 is similar to the D1 but collected independently. All three model's performances are depicted in Table \ref{tab:dataset2}. Even though the performance of both Qu's model and ResUnet decreases on D2, we do not observe any reduction in the HoVer-Net model's performance. It highlights that the HoVer-Net model trained using point annotations has a better generalization capability in comparison to the two baseline approaches in terms of the detection metrics.    

\begin{table}[t]
  \centering 
  \caption{Comparative Analysis of Generalization Capability on D2}
  \begin{tabular}{llll}
  \toprule
    \textbf{Method} & \textbf{Precision} & \textbf{Recall} & \textbf{F1} \\
    \midrule
    HoVer-Net Ent & 0.918 & 0.892 & 0.904 \\ 
    Res-Unet34 & 0.9054 & 0.7132 & 0.7981 \\ 
    Qu's approach & 0.8471 & 0.7969 & 0.8122\\ 
    \bottomrule
  \end{tabular}
  \label{tab:dataset2} 
\end{table}

\subsection{Results on Multi-Organ (MO) dataset:} 

We compare the performance of the HoVer-Net model trained using point annotations with the Qu's model over the MO dataset. We follow the same modeling set-up used in \cite{qu2019weakly} and compare the performance of our approach with the numbers reported in their paper.  To calculate the performance of the HoVer-Net model in terms of the prediction metrics, we set $r_{nuc} = 11$ as suggested by \cite{qu2019weakly}. As shown in Table \ref{tab:MO}, the HoVer-Net model outperforms Qu's model in terms of precision and F1 score. In addition, the results show that adapting the entropy minimization while training the HoVer-Net model on the MO dataset only results in a small improvement in the recall metric. 

\begin{table}[t]
  \centering 
  \caption{Comparative experiments on the Results on MO dataset}
  \begin{tabular}{llll}
  \toprule
    \textbf{Method} & \textbf{Precision} & \textbf{Recall} & \textbf{F1} \\
    \midrule
    HoVer-Net & 0.8886 & 0.8310 & 0.8559 \\ 
    HoVer-Net Ent& 0.8847 & 0.8328& 0.8548 \\
    Qu's approach & 0.8420 & 0.8665 & 0.8541\\ 
    \bottomrule
  \end{tabular}
  \label{tab:MO} 
\end{table}

%% file: Discussion.tex
%problem and challenges
Developing an automated approach for localizing and counting the various cell types in 3D immunofluorescent images is a challenging task. The main challenges are 1) they have various z-axis, 2) they contain many overlapping nuclei that make the counting task difficult, and 3) the pixel-level annotation of cells in 3D images is costly and time-consuming.  

% our approach
To alleviate the challenges mentioned above, we adapted a weakly supervised learning approach to train the HoVer-Net model using point labels for nuclei segmentation in z-stack immunofluorescent images. To convert the 3D images to 2D (z-stack) immunofluorescent images, we used a Maximum Intensity Projection (MIP) algorithm. The reason for adopting the HoVer-Net model is its strong nuclei instance detection capability, making it a good fit for detecting nuclei in z-stack images as they contain touching and sometimes overlapping nuclei. However, to train the HoVer-Net model, we needed to create binary masks which could be generated using pixel-level annotations. As previously mentioned, generating pixel-level annotations is expensive and time-consuming. By allowing clinicians to generate point-level annotations instead (i.e. marking only the center of the nuclei), we reduce the annotation burden and increase the likelihood of this method being adopted. Herein we utilized the cluster label approach proposed in \cite{qu2019weakly} to generate binary masks using point labels to train the HoVer-Net model.
However, the generated pixel-level masks are not perfect, and they contain unlabeled areas, especially around the boundaries of the nuclei. Therefore, we adapted the semi-supervised learning approach and used entropy minimization loss in these unlabeled areas to improve the training process. The reason for training the model on unlabeled areas is that they may contain important information. In addition, applying the entropy minimization loss function encourages the model to output confident predictions on unlabeled areas. 

% findings
In experimental results, we compare the performance of the HoVer-Net model trained using point annotations with baseline methods for detecting nuclei on the z-stack immunofluorescent images. The comparative analysis indicates that the HoVer-Net model outperforms the baseline methods. In addition, we show that using the entropy minimization loss in these areas can further improve the recall metric. More importantly, as achieving good performance on the unseen data is the ultimate goal of designing the automatic approach, we show that the HoVer-Net model has a better generalization capability in terms of detection metrics in comparison to the baseline methods. Finally, the results of the experiments on the public dataset show that the HoVer-Net method achieves comparable performance to Qu's approach. 

% future works
In our future work, we will explore the simultaneous segmentation and classification of various cells in 3D immunofluorescent images that are highly imbalanced.

\paragraph{Limitations} In this work, we adapted a weakly supervised learning approach to train the HoVer-Net model using point labels for nuclei segmentation in z-stack immunofluorescent images and compare it to baseline methods. We showed that the HoVer-Net model using weakly supervised learning outperforms the current baseline methods in terms of detection metrics. However, there are multiple hyper-parameters in the HoVer-Net model and baseline methods. We only manually tuned the value of the hyper-parameters corresponding to the loss functions' weights and the number of the epochs in the HoVer-Net model. All other hyper-parameters of the used methods are set to predefined values. We did not use any sophisticated hyper-parameter tuning approach to search for hyperparameters values for any of the approaches. However, there is a possibility to obtain different results by using sophisticated hyper-parameter tuning. Also, while the range of z-slices in the cardiovascular datasets was varied, the samples were sectioned at $10\mu m$. Given that the images were collapsed into a 2D image using MIP, we foresee that this method would become limited if thicker tissue sections were analyzed. Further work into image sub-sampling in the z-plane will need to be conducted to resolve this issue.

\paragraph{Acknowledgements}

This work was supported under grants NHLBI R01 HL156849-01, NHLBI R01 HL155165-01, NHLBI R01 HL156849-01, and NHLBI R01 HL141425-01. Also, this work was provided partially by a grant to the integrated Translational Health Research Institute (iTHRIV) with funding support from National Center for Advancing Translational Sciences (NCATS) UL1 TR003015.